\documentclass[a4paper,fleqn]{cas-dc}

\usepackage[authoryear,longnamesfirst]{natbib}
\usepackage{graphicx}
\usepackage{tabularx}

\def\tsc#1{\csdef{#1}{\textsc{\lowercase{#1}}\xspace}}
\tsc{WGM}
\tsc{QE}
\begin{document}
\let\WriteBookmarks\relax
\def\floatpagepagefraction{1}
\def\textpagefraction{.001}

\shorttitle{}    

\shortauthors{}  

\title [mode = title]{VPTracker: Global Vision-Language Tracking via Visual Prompt}  
\tnotemark[1] 


\tnotetext[1]{The paper is under consideration at Pattern Recognition Letters.} 


% \author[1]{}%[<options>]

% % Corresponding author indication
% \cormark[1]

% % Footnote of the first author

% % Email id of the first author
% \ead{}

% % URL of the first author

% % Credit authorship
% % eg: \credit{Conceptualization of this study, Methodology, Software}

% % Address/affiliation
% \affiliation[1]{organization={},
% %          citysep={}, % Uncomment if no comma needed between city and postcode

% \author[2]{}%[]

% % Footnote of the second author

% % Email id of the second author
% \ead{}

% % URL of the second author

% % Credit authorship

% % Address/affiliation
% \affiliation[2]{organization={},
% %          citysep={}, % Uncomment if no comma needed between city and postcode

% % Corresponding author text
% \cortext[1]{Corresponding author}

% % Footnote text

% 作者列表
\author[dsae]{Jingchao Wang}
\ead{jcwang@stu.ecnu.edu.cn}
\credit{Writing - Original Draft, Writing - Review & Editing, Investigatio, Conceptualizationn, Methodology, Data curation, Validation, Software}
\address[dsae]{School of Data Science and Engineering, East China Normal University, Shanghai 200062, China}

\author[dsae]{Kaiwen Zhou}
\ead{51275903051@stu.ecnu.edu.cn}
\credit{Writing - Original Draft, Investigation, Conceptualization, Methodology, Data curation, Validation, Software}

\author[west]{Zhijian Wu}
\ead{wuzhijian@westlake.edu.cn}
\address[west]{Medical Artificial Intelligence Laboratory, Westlake University, Hangzhou 310024, China}
\credit{Writing - Review & Editing, Investigation, Conceptualization, Methodology, Validation}

\author[dsae]{Kunhua Ji}
\ead{72285900034@stu.ecnu.edu.cn}
\credit{Writing - Review & Editing, Investigation, Conceptualization, Methodology, Validation}

\author[dsae]{Dingjiang Huang}
\ead{djhuang@dase.ecnu.edu.cn}
\cormark[1]
\credit{Writing - Review & Editing, Investigation, Conceptualization, Supervision, Funding acquisition}

\author[west]{Yefeng Zheng}
\ead{zhengyefeng@westlake.edu.cn}
\credit{Writing - Review & Editing, Investigation, Conceptualization, Supervision}

% 通讯作者脚注标记

\cortext[cor]{Corresponding author}
% 单位地址定义

\newcommand{\mymethod}{\textbf{VPTracker}}
\begin{abstract}
Vision-Language Tracking aims to continuously localize objects described by a visual template and a language description. Existing methods, however, are typically limited to local search, making them prone to failures under viewpoint changes, occlusions, and rapid target movements. In this work, we introduce the first global tracking framework based on Multimodal Large Language Models (VPTracker), exploiting their powerful semantic reasoning to locate targets across the entire image space. While global search improves robustness and reduces drift, it also introduces distractions from visually or semantically similar objects. To address this, we propose a location-aware visual prompting mechanism that incorporates spatial priors into the MLLM. Specifically, we construct a region-level prompt based on the target's previous location, enabling the model to prioritize region-level recognition and resort to global inference only when necessary. This design retains the advantages of global tracking while effectively suppressing interference from distracting visual content. Extensive experiments show that our approach significantly enhances tracking stability and target disambiguation under challenging scenarios, opening a new avenue for integrating MLLMs into visual tracking. Code and models are available at \url{https://github.com/jcwang0602/VPTracker}.
\end{abstract}


% \begin{highlights}
% \item First MLLM-based global tracker enabling full-image target search.

% \item Location-aware visual prompting reduces distractors, improves accuracy.

% \item Outperforms SOTA with stronger robustness and long-term stability.
% \end{highlights}

%\nocite{*}

% Keywords

\maketitle

\section{Introduction}
\label{sec:intro}

Visual object tracking (VOT) is a fundamental problem in computer vision, supporting a wide range of downstream applications such as intelligent surveillance, human–robot interaction, augmented reality, and autonomous systems~\cite{10149530,ZHAO202679,GE202615,XING202559}. Traditional visual trackers primarily rely on an initial visual template and perform frame-by-frame localization using appearance similarity~\cite{ye2022ostrack}. Although such methods have achieved strong short-term performance, they are inherently constrained by the limited representational capacity of a single visual template. As a result, their robustness deteriorates in complex scenarios involving drastic appearance changes, heavy occlusions, large camera motion, or long-term absence.

To overcome the ambiguity of purely visual cues, vision-language tracking has recently emerged as a promising direction~\cite{11209477}. 
By incorporating natural language descriptions, trackers can leverage high-level semantic priors to better distinguish objects with similar appearance and maintain identity consistency over time. 
This paradigm has shown particular advantages in multi-object scenes and cases where visual cues alone are insufficient~\cite{wang2021towards}. 
However, existing most vision-language trackers still predominantly adopt local search strategies, restricting the search region to a small neighborhood around the previous target location. 
While efficient, this localized assumption makes the tracker vulnerable to drift: once the target moves rapidly, undergoes abrupt viewpoint changes, or becomes temporarily occluded, the tracker may fail to rediscover it.

Recent advancements in Multimodal Large Language Models (MLLMs) have opened new possibilities for rethinking the tracking pipeline. 
MLLMs exhibit remarkable capabilities in global scene understanding, cross-modal reasoning~\cite{bai2025qwen3}. 
These strengths suggest that MLLMs could fundamentally reshape long-term tracking, allowing the model to perform global search over the entire image rather than relying on local heuristics. 
In this paper, we present the first global vision–language tracking framework powered by Multimodal Large Language Models (MLLMs), termed {\mymethod}. Leveraging the strong semantic reasoning capabilities of MLLMs, our approach can localize the target across the entire image space.
Despite their impressive semantic understanding, we observe that large-scale language models inherently lack temporal modeling and spatial continuity awareness—two key elements required for reliable visual tracking.
To overcome this limitation, we introduce a location-aware visual prompting mechanism that injects spatial priors directly into the MLLM’s inference process. 
Specifically, we construct a region-level prompt based on the target’s previous location, guiding the model to focus on the most probable search area while still maintaining the ability to perform global reasoning when necessary. 
This design preserves the robustness of global search while substantially reducing false positives caused by visually or semantically similar distractors.

The experimental results on the commonly used visual language tracking datasets TNL2K~\cite{wang2021towards}, and TNLLT~\cite{wang2025reasoningtrack} demonstrate that our method has achieved significant improvements in robustness, target re-identification, and long-term stability. 
More importantly, {\mymethod} introduces a new perspective of integrating MLLM into vision-language tracking tasks, indicating that future tracking will not only be driven by appearance similarity but also by advanced cross-modal reasoning. 
In summary, the main contributions of this work are threefold:

\begin{itemize}

\item We propose the first global vision–language tracking framework built on Multimodal Large Language Models (MLLMs), enabling global target localization across the entire image space rather than relying on traditional local search heuristics. This introduces a new paradigm for long-term tracking driven by cross-modal semantic reasoning.

\item We identify the lack of temporal modeling and spatial continuity in MLLMs and address it with a location-aware visual prompting mechanism. By injecting region-level spatial priors derived from the target’s previous state, our method significantly reduces false positives and enhances target recognition while preserving global reasoning capabilities.

\item Extensive experiments on TNL2K and TNLLT demonstrate that our approach achieves superior robustness, improved recognition capability, and strong long-term stability, establishing a new direction for integrating MLLMs into vision–language tracking.
\end{itemize}

\section{Related Work}
\label{sec:relatedwork}
\subsection{Object Tracking}

Visual object tracking (VOT) has been extensively studied over the past decade, with research evolving from correlation filter–based methods to modern deep learning–based frameworks~\cite{ye2022ostrack}. Early trackers such as KCF~\cite{henriques2014high} and ECO~\cite{danelljan2017eco} model target appearance through handcrafted features and efficient frequency-domain operations. With the success of deep convolutional networks, Siamese-based trackers (e.g., SiamFC~\cite{bertinetto2016fully}, SiamRPN~\cite{li2018high}, SiamMask~\cite{hu2023siammask}) became dominant due to their strong template-matching capability and real-time performance. Transformer-based trackers (e.g., TransT~\cite{chen2021TransT}, STARK~\cite{tang2022SFTransT}, OSTrack~\cite{ye2022ostrack}) further improve robustness by modeling long-range dependencies with self-attention mechanisms.

The introduction of natural language into visual tracking provides richer semantic information beyond appearance cues. Early works such as MANet and ReferFormer incorporate text embeddings into the tracking pipeline to better distinguish visually similar objects. LST, TNL2K-trackers, and subsequent models integrate joint vision-language encoders (e.g., CLIP) to enhance target localization through semantic alignment. These approaches demonstrate the advantages of textual descriptions in improving identity consistency.

However, whether in visual tracking or vision–language tracking, most existing methods still follow a local search paradigm, where the search region is restricted to a fixed-size window centered around the last known target position.
When the target undergoes large displacement or disappears due to occlusion, the tracker often fails to re-detect it. Moreover, most existing methods rely on shallow fusion of linguistic and visual features, restricting the model’s ability to reason globally about object semantics, inter-object relationships, and scene-level context. As a result, they struggle in complex environments with distractors that share similar descriptions or visual patterns.

\subsection{Multimodal Large Language Models}

Multimodal Large Language Models (MLLMs), such as GPT-4V~\cite{achiam2023gpt}, LLaVA~\cite{li2024llava}, Qwen-VL\cite{bai2025qwen3}, and InternVL~\cite{wang2025internvl3}, have exhibited strong capabilities in high-level scene understanding, cross-modal reasoning, and open-vocabulary recognition~\cite{wang2025unlocking}. Unlike traditional vision-language models that output fixed embeddings, MLLMs are able to perform explicit reasoning, generate spatial descriptions, and interpret complex relationships between objects. These properties make them promising for challenging video tasks such as grounding, captioning, and instruction-following.

Nevertheless, applying MLLMs directly to visual tracking remains underexplored. Existing works primarily focus on isolated tasks such as image-level grounding or referring expression comprehension, without considering temporal consistency, spatial priors, or the recursive nature of tracking. Moreover, MLLMs often favor global but semantically-driven interpretation, which can lead to ambiguous predictions when visually similar objects appear in the scene. This reveals a fundamental gap between the global semantic reasoning strengths of MLLMs and the precise, temporally grounded localization required for tracking.

\section{Method}
\label{sec:method}

\begin{figure*}[t]
    \centering
    \includegraphics[width=1\linewidth]{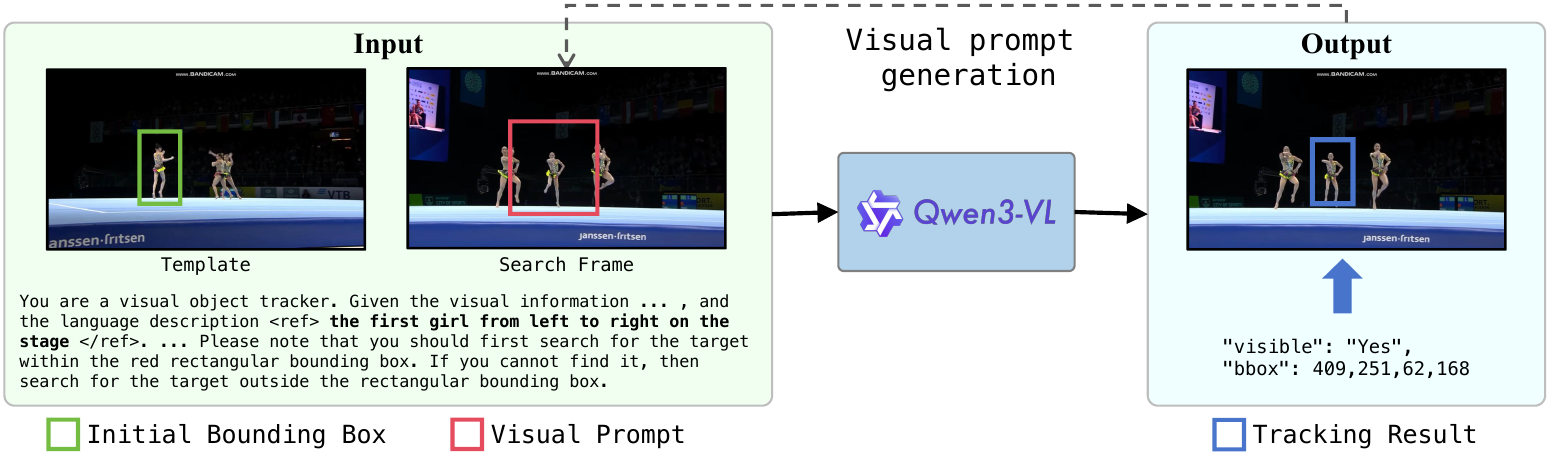}
    \caption{An overview of our proposed VPTracker. }
    \label{fig:method}
\end{figure*}

\begin{figure*}[t]
    \centering
    \includegraphics[width=1\linewidth]{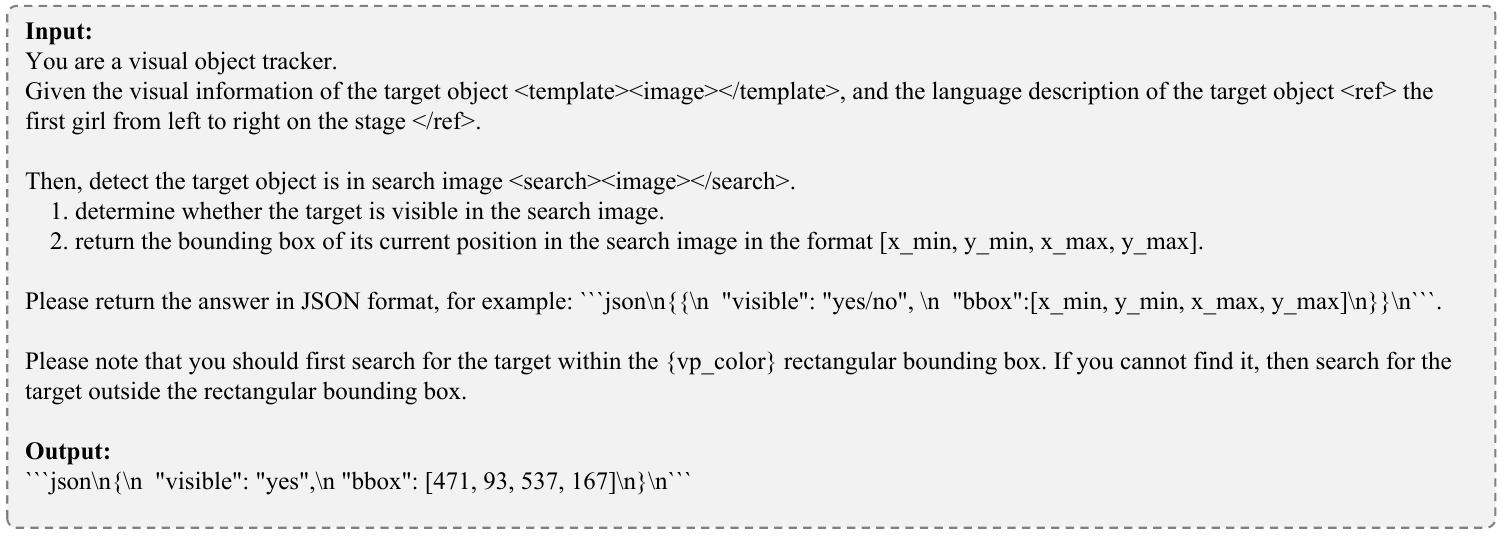}
    \caption{Prompt used for Vision-Language Tracking.}
    \label{fig:prompt}
\end{figure*}

\subsection{Problem Formulation}
Visual-language tracking aims to localize an object in subsequent video frames based on the given bounding box and the corresponding language description in the first frame. 
Most existing trackers rely on the assumption that the target moves continuously and smoothly throughout the video. 
A dominant paradigm is to restrict the search region in the current frame using the target’s predicted position and size from the previous frame, allowing the model to perform localization only within a limited area. The overall pipeline can be summarized as follows:
\begin{align}
& T = \text{Crop}(I_1, B_1) \\
& S_t = \text{Crop}(I_t, B_{t-1}), \quad t=2,\dots,N \\
& B_t = M(T, L, S_t)
\end{align}
where $I_t$ denotes the \(t\)-th video frame; 
$B_t$ is the predicted bounding box of the target in frame $I_t$; 
$L$ is the natural language description of the target; 
$\text{Crop}(I_t, B_{t-1})$ denotes the region cropped from the full image $I_t$ based on the previous bounding box $B_{t-1}$, after enlarging it by a certain scale;
and $M(T, L, S_t)$ denotes the localization operator that outputs the predicted bounding box $B_t$.
This typical paradigm of cropping the search region, while avoiding excessive visual information interference and reducing the tracking difficulty for the model, also leads to target loss when encountering scenarios such as excessively fast target movement, viewpoint switching, or target occlusion.

\subsection{Global Tracking via MLLMs}

To address the aforementioned issues, this paper proposes a global tracking method based on MLLM. Leveraging MLLM's powerful reasoning capabilities, it searches for targets across the entire video frame given a visual template and corresponding linguistic description, and provides bounding boxes for the detected targets.
As illustrated in Fig.~\ref{fig:method}, our approach relies exclusively on a single multimodal large language model, without requiring any additional components. The prompt employed in our method is presented in Fig.~\ref{fig:prompt}.

Specifically, we first crop a visual template $T$ from the initial bounding box in the first frame. 
Taking the second frame as an example, we draw a rectangle on $I_2$ based on $B_1$ to form a visual prompt, while the drawing details will be introduced in the next subsection. 
Then, the visual template $T$, the current frame with the visual prompt $I_t'$, and the instruction text $L'$ containing the language description are fed into the MLLM to obtain the tracking result.
\begin{align}
& T = \text{Crop}(I_1, B_1) \\
& I_t'= \text{VP}(I_t, B_{t-1}), \quad t=2,\dots,N \\
& B_t = MLLM(T, L', I_t')
\end{align}
where $\text{VP}(I_t, B_{t-1})$ denotes drawing the visual prompt on the current frame $I_t$ based on $B_{t-1}$. However, compared with local tracking paradigms, the global tracking paradigm can more effectively re-acquire the target after rapid motion or temporary disappearance. Nevertheless, the enlarged search space inevitably introduces more irrelevant visual information, which increases the difficulty of accurate tracking.

\subsection{Visual Prompt based on Location-aware}

To mitigate the performance degradation caused by visual distractions in global tracking, we introduce a location-aware visual prompting mechanism. The core idea is consistent with local tracking methods, which assume that the target's motion trajectory is relatively smooth; thus, the target's position in frame $t+1$ is typically close to its position in frame $t$. Unlike conventional local tracking approaches that crop a small search region---thereby discarding information outside the cropped region---we avoid restricting the model's field of view.

Instead, we draw a rectangular box directly on the full search frame as a visual prompt, encouraging the model to prioritize searching within this region. When the target moves out of the prompted region, the model then falls back to a full-frame global search. This design preserves the advantages of global perception while effectively suppressing interference from large-scale background content.

To prevent the model from overfitting to the presence of bounding boxes, we additionally include a proportion of training samples where the visual prompt is drawn but the target lies outside the prompted region. This prevents the model from becoming overly reliant on the boxes themselves. However, during inference, we strictly draw bounding boxes based on the model's predictions.

\section{Experiment}
\label{sec:exp}

\subsection{Implementation Details}

\paragraph{Dataset}

We train our model using a mixed training set composed of TNL2K and TNLLT. The TNL2K training split contains 1,300 videos, while the TNLLT training split includes 150 long video sequences. We sample the data using a 7:3 ratio, resulting in a total of one million training samples. Each sample consists of a visual template, a search frame, and the corresponding language description.

\paragraph{Training and Inference}
We adopted Qwen3-VL-4B as our base model. 
All experiments are conducted on an 8 NVIDIA H200 GPU. 
We use the Adamw optimizer with an initial learning rate of 2e-5 and a batch size of 128. 
During training, we fine-tune all layers of Qwen3-VL including vision encoder, alignment layer and llm.
% 整个训练过程大约需要 8 个小时左右。
The entire training process takes approximately 8 hours.
During inference, we adopt the default decoding and reasoning configuration of Qwen3-VL. 
Each frame is processed independently by feeding the full-resolution image into the model.

\subsection{Comparison with State-of-the-art Methods}

We test VPTracker on the large-scale test datasets of TNL2K and TNLLT, and conduct a comprehensive comparison with state-of-the-art methods..

\paragraph{TNLLT} The TNLLT dataset is a recently proposed benchmark for long-term vision–language tracking, designed to assess a model’s capability to localize target objects over extended video sequences using natural language descriptions. Its test set consists of 50 long and diverse video sequences.
To better assess robustness, TNLLT further provides 15 challenging attributes, such as occlusion, deformation, motion blur, illumination variation, fast motion, low resolution, out-of-view, and background clutter. These rich annotations make TNLLT a comprehensive benchmark for evaluating long-term temporal modeling, and cross-modal reasoning in vision–language tracking.
We compare {\mymethod} with existing state-of-the-art trackers in Tab.\ref{tab:tnllt}.
Compared to the state-of-the-art ReasoningTrack, which uses MLLM to update the linguistic description of the object during tracking, we achieve significant improvements over prior methods.

\begin{figure*}[t]
    \centering
    \includegraphics[width=1.0\linewidth]{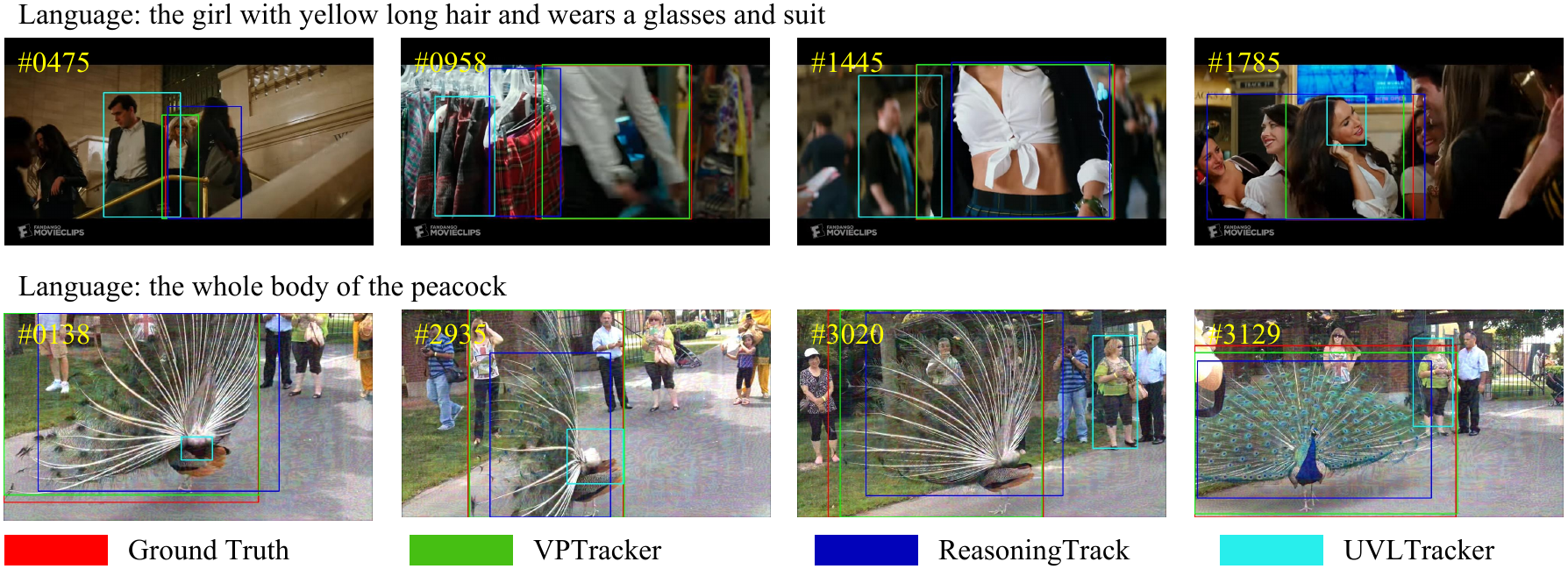}
    \caption{Visual comparison of tracking results with and without Visual Prompt}
    \label{fig:result}
\end{figure*}

\begin{figure*}[t]
    \centering
    \includegraphics[width=1.0\linewidth]{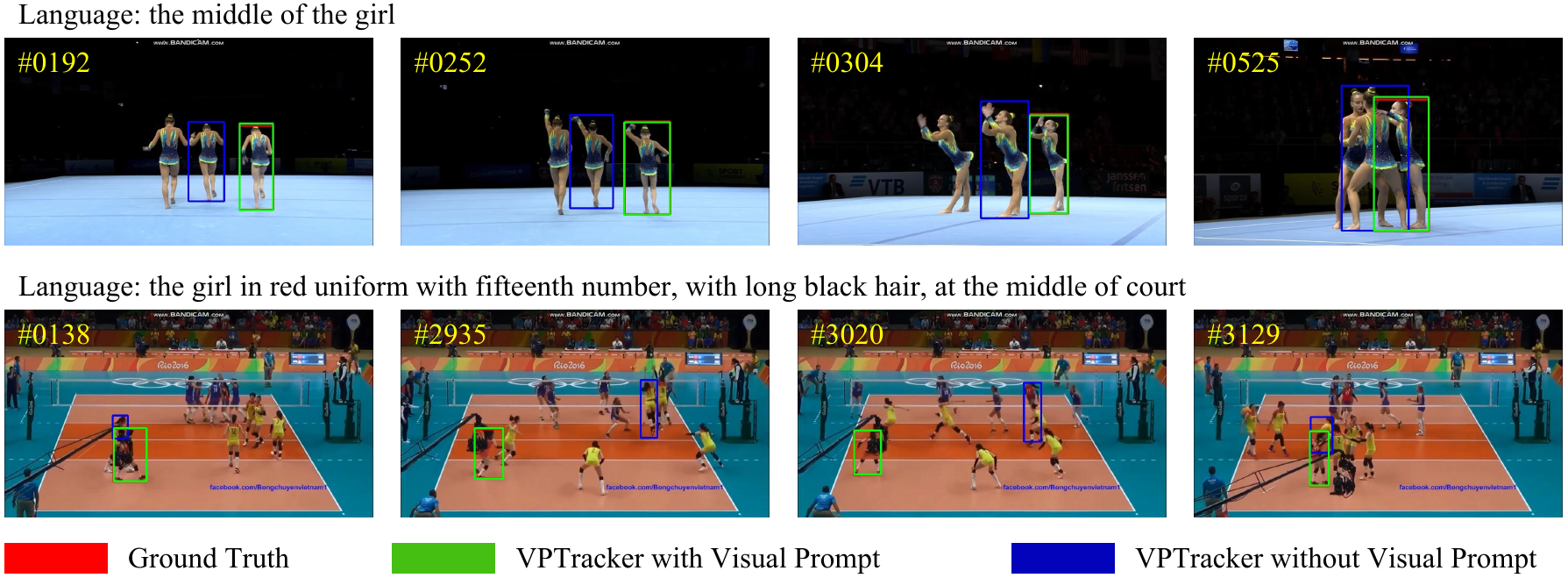}
    \caption{Visual comparison of tracking results with and without Visual Prompt}
    \label{fig:ablation}
\end{figure*}

\begin{table}[t]
\centering
\caption{Overall tracking performance on TNLLT dataset. The best results are highlighted in bold.}
% \begin{tabularx}{\linewidth}{X|c|c|ccc}
\resizebox{\linewidth}{!}{
\begin{tabular}{l|c|c|ccc}
\toprule
\multirow{2}{*}{Method}   & \multirow{2}{*}{Source} & \multirow{2}{*}{Type} & \multicolumn{3}{c}{TNLLT} \\
  & &           & $SR$   & $PR$   & $NPR$   \\
\hline
OSTrack~\cite{ye2022ostrack}          & ECCV22   & BB        & 52.1   & 57.3   & 63.6    \\
MixFormer~\cite{cui2022mixformer}     & CVPR22   & BB        & 56.0   & 61.1   & 67.8    \\
AiATrack~\cite{gao2022aiatrack}       & ECCV22   & BB        & 53.0   & 56.9   & 63.5    \\
JointNLT~\cite{zhou2023joint}         & CVPR23   & NL        & 45.0   & 47.9   & 55.5    \\
JointNLT~\cite{zhou2023joint}         & CVPR23   & BL        & 45.4   & 48.2   & 55.9    \\
All-in-one~\cite{zhang2023all}        & MM23     & BL        & 49.7   & 54.8   & 60.6    \\
MMTrack~\cite{zheng2023toward}        & TCSVT23  & BL        & 55.8   & 61.8   & 67.7    \\
CiteTrack~\cite{citetracker}          & ICCV23   & BB        & 50.3   & 55.3   & 60.8    \\
ROMTrack~\cite{cai2023robust}         & ICCV23   & BB        & 49.0   & 54.8   & 59.9    \\
GRM~\cite{gao2023generalized}         & CVPR23   & BB        & 50.0   & 55.4   & 61.3    \\
ODTrack~\cite{zheng2024odtrack}       & AAAI24   & BB        & 54.4   & 61.3   & 65.9    \\
EVPTrack~\cite{shi2024evptrack}       & AAAI24   & BB        & 54.4   & 61.3   & 65.8    \\
UVLTrack~\cite{ma2024unifying}        & AAAI24   & BB        & 53.1   & 58.1   & 64.1    \\
AQATrack~\cite{xie2024autoregressive} & CVPR24   & BB        & 55.1   & 61.5   & 66.5    \\
UVLTrack~\cite{ma2024unifying}        & AAAI24   & NL        & 51.9   & 57.4   & 63.1    \\
UVLTrack~\cite{ma2024unifying}        & AAAI24   & BL        & 54.4   & 60.3   & 66.2    \\
SUTrack~\cite{sutrack}                & AAAI25   & BL        & 50.8   & 55.4   & 60.0    \\
LMTrack~\cite{xu2025less}             & AAAI25   & BB        & 48.7   & 53.0   & 59.8    \\
CTVLT~\cite{feng2025enhancing}        & ICASSP25 & BL        & 60.9   & 69.5   & 73.6    \\
% DUTrack~\cite{Li2025dutrack}          & CVPR 2025   & BL        & 62.8   & 72.5   & 75.8    \\
% ReasoningTrack~\cite{wang2025reasoningtrack}  & Arxiv 2025           & BL        & 63.9   & \textbf{74.1}   & 77.0    \\
VPTracker    & Ours   & BL  & \textbf{62.5}  & \textbf{72.9}  & \textbf{73.8}     \\ 
\bottomrule
% \end{tabularx}
\end{tabular}
}
\label{tab:tnllt}
\end{table}

\paragraph{TNL2K} The TNL2K dataset is a large dataset used for training and evaluating Vision-Language tracking. It contains 2000 video sequences and 663 words, of which 1300 videos are used for training and 700 videos are used for testing. It introduces two new challenges: adversarial samples and modality switches. We compare {\mymethod} with existing advanced trackers in Tab.\ref{tab:tnl2k}. we achieve significant improvements over prior methods.

\begin{table}[ht]
\centering
\caption{Overall tracking performance on TNL2K dataset. The best results are highlighted in bold.}
\resizebox{\linewidth}{!}{
\begin{tabular}{l|c|c|ccc}
\toprule
\multirow{2}{*}{Method}      & \multirow{2}{*}{Source} & \multirow{2}{*}{Type} & \multicolumn{3}{c}{TNL2K}\\
     & &           & $AUC$       & $PR$        & $NPR$ \\
\hline
TrDiMP~\cite{wang2021transformer}       & CVPR21    & BB   & -          & -          & -          \\
TransT~\cite{tang2022SFTransT}          & CVPR21    & BB   & 50.7       & 51.7       & 57.1       \\
AutoMatch~\cite{zhang2021learn}         & AAAI21    & BB   & 47.2       & 43.5       & -          \\
SNLT~\cite{feng2021siamese}             & CVPR21    & BL   & 27.6       & 41.9       & -          \\
% CapsuleTNL~\cite{ma2021capsule}         & MM21 & BL   & -          & -          & -          \\
VLTTT~\cite{guo2022divert}              & NIPS22 & BL   & 53.1       & 53.3       & -          \\
CTRNLT~\cite{li2022cross}               & CVPR22    & BL   & 44.0       & 45.0       & 52.0       \\
JointNLT~\cite{zhou2023joint}           & CVPR23    & BL   & 56.9       & 58.1       & 73.6       \\
OneTrack~\cite{hong2024onetracker}      & CVPR24    & BL   & 58.0       & 59.1       & -          \\
QueryNLT~\cite{shao2024context}         & CVPR24    & BL   & 57.8       & 58.7       & 75.6       \\
MAVLT~\cite{shi2025mamba}               & TCSVT25 & BL  & 63.1  &  66.7 & - \\
ProVLT~\cite{10947500}                  & TCSVT25 & BL  & 59.8  & 61.8  & - \\
TemTrack~\cite{xie2025robust}           & AAAI25 & BB & 58.8 & - & - \\
VPTracker                    & Ours        & BL   & \textbf{64.9}       & \textbf{71.2}       & \textbf{80.2}       \\ 
\bottomrule
% \end{tabularx}
\end{tabular}
}
\label{tab:tnl2k}
\end{table}

\subsection{Qualitative Comparison}
To better demonstrate the superiority of our method, we compared its tracking results with those of two state-of-the-art approaches: ReasoningTrack and UVLTrack. As shown in Fig.~\ref{fig:result}, our method achieves more accurate tracking performance when dealing with scenes featuring significant changes in target appearance and numerous distracting factors.

\subsection{Ablation study}
To validate the effectiveness of our proposed location-aware visual prompting mechanism, we conduct comparisons on the TNL2K dataset with and without the visual prompt. As shown in Tab.~\ref{tab:ablation}, incorporating the visual prompting mechanism leads to consistent and notable improvements across metrics such as AUC, PR, and NPR.
To more clearly demonstrate the effectiveness of our proposed VP mechanism, we present a visual comparison of the tracking results in Fig.~\ref{fig:ablation}. Since the textual description of the target may only be applicable to the first frame, global tracking without the position-aware visual prompt can lead to incorrect prediction when visually similar objects appear in the scene.

\begin{table}[ht]
\centering
\caption{Overall tracking performance on TNL2K dataset.}
\begin{tabular}{l|ccc}
\toprule
\multirow{2}{*}{Method}     & \multicolumn{3}{c}{TNL2K}\\
        & $AUC$  & $PR$   & $NPR$ \\
\hline
w. VP   & 64.9   & 71.2   & 80.2  \\
w/o VP  & 64.3   & 70.1   & 79.2  \\ 
\bottomrule
\end{tabular}
\label{tab:ablation}
\end{table}

% \vspace{-3mm}

\section{Conclusion}
In this paper, we presented VPTracker, a global vision–language tracking framework built upon a Multimodal Large Language Model. By introducing a simple yet effective location-aware visual prompting mechanism, our method successfully injects spatial priors into the MLLM, enabling it to suppress distractors while maintaining global perception. Experiments on TNL2K and TNLLT show clear improvements in robustness and long-term stability. These results demonstrate that MLLMs, when properly guided, can serve as a strong backbone for next-generation global tracking systems.

Although VPTracker demonstrates competitive performance in vision–language tracking, it represents only an initial exploration, and we have not yet incorporated more sophisticated mechanisms to further boost its capability. We believe that MLLMs hold substantial untapped potential for the tracking community, and we hope this work encourages more researchers to explore this promising direction.

\section*{Acknowledgments}
This work was partially supported by the National Natural Science Foundation of China under Grant 62072185.


\bibliographystyle{unsrt}
% \bibliographystyle{cas-model2-names}

\bibliography{cas-refs}

\begin{thebibliography}{10}

\bibitem{10149530}
Huanlong Zhang, Jingchao Wang, Jianwei Zhang, Tianzhu Zhang, and Bineng Zhong.
\newblock One-stream vision-language memory network for object tracking.
\newblock {\em IEEE Transactions on Multimedia}, 26:1720--1730, 2024.

\bibitem{ZHAO202679}
Junzhe Zhao, Jintao Su, Ye~Liu, Jun Liu, and Miaohui Wang.
\newblock Dptracker: Dynamic prompter for rgb-d tracking.
\newblock {\em Pattern Recognition Letters}, 204:79--85, 2026.

\bibitem{GE202615}
Jingzhan Ge and Kamran Mohseni.
\newblock Input-augmented taylor expansion (iate) kalman filter for robust visual object tracking.
\newblock {\em Pattern Recognition Letters}, 202:15--21, 2026.

\bibitem{XING202559}
Haijiao Xing, Wei Wei, Lei Zhang, and Chen Ding.
\newblock A visual prompt learning network for hyperspectral object tracking.
\newblock {\em Pattern Recognition Letters}, 196:59--65, 2025.

\bibitem{ye2022ostrack}
Botao Ye, Hong Chang, Bingpeng Ma, Shiguang Shan, and Xilin Chen.
\newblock Joint feature learning and relation modeling for tracking: A one-stream framework.
\newblock In {\em ECCV}, 2022.

\bibitem{11209477}
Jingchao Wang, Zhijian Wu, Wenlong Zhang, Wenhui Liu, Jianwei Zhang, and Dingjiang Huang.
\newblock Overcoming feature contamination by unidirectional information modeling for vision-language tracking.
\newblock In {\em 2025 IEEE International Conference on Multimedia and Expo (ICME)}, pages 1--6, 2025.

\bibitem{wang2021towards}
Xiao Wang, Xiujun Shu, Zhipeng Zhang, Bo~Jiang, Yaowei Wang, Yonghong Tian, and Feng Wu.
\newblock Towards more flexible and accurate object tracking with natural language: Algorithms and benchmark.
\newblock In {\em Proceedings of the IEEE/CVF conference on computer vision and pattern recognition}, pages 13763--13773, 2021.

\bibitem{bai2025qwen3}
Shuai Bai, Yuxuan Cai, Ruizhe Chen, Keqin Chen, Xionghui Chen, Zesen Cheng, Lianghao Deng, Wei Ding, Chang Gao, Chunjiang Ge, et~al.
\newblock Qwen3-vl technical report.
\newblock {\em arXiv preprint arXiv:2511.21631}, 2025.

\bibitem{wang2025reasoningtrack}
Xiao Wang, Liye Jin, Xufeng Lou, Shiao Wang, Lan Chen, Bo~Jiang, and Zhipeng Zhang.
\newblock Reasoningtrack: Chain-of-thought reasoning for long-term vision-language tracking.
\newblock {\em arXiv preprint arXiv:2508.05221}, 2025.

\bibitem{henriques2014high}
Jo{\~a}o~F Henriques, Rui Caseiro, Pedro Martins, and Jorge Batista.
\newblock High-speed tracking with kernelized correlation filters.
\newblock {\em IEEE transactions on pattern analysis and machine intelligence}, 37(3):583--596, 2014.

\bibitem{danelljan2017eco}
Martin Danelljan, Goutam Bhat, Fahad Shahbaz~Khan, and Michael Felsberg.
\newblock Eco: Efficient convolution operators for tracking.
\newblock In {\em Proceedings of the IEEE conference on computer vision and pattern recognition}, pages 6638--6646, 2017.

\bibitem{bertinetto2016fully}
Luca Bertinetto, Jack Valmadre, Joao~F Henriques, Andrea Vedaldi, and Philip~HS Torr.
\newblock Fully-convolutional siamese networks for object tracking.
\newblock In {\em European conference on computer vision}, pages 850--865. Springer, 2016.

\bibitem{li2018high}
Bo~Li, Junjie Yan, Wei Wu, Zheng Zhu, and Xiaolin Hu.
\newblock High performance visual tracking with siamese region proposal network.
\newblock In {\em Proceedings of the IEEE conference on computer vision and pattern recognition}, pages 8971--8980, 2018.

\bibitem{hu2023siammask}
Weiming Hu, Qiang Wang, Li~Zhang, Luca Bertinetto, and Philip~HS Torr.
\newblock Siammask: A framework for fast online object tracking and segmentation.
\newblock {\em IEEE Transactions on Pattern Analysis and Machine Intelligence}, 45(3):3072--3089, 2023.

\bibitem{chen2021TransT}
Xin Chen, Bin Yan, Jiawen Zhu, Dong Wang, Xiaoyun Yang, and Huchuan Lu.
\newblock Transformer tracking.
\newblock In {\em Proceedings of the IEEE/CVF Conference on Computer Vision and Pattern Recognition}, pages 8126--8135, 2021.

\bibitem{tang2022SFTransT}
Chuanming Tang, Xiao Wang, Yuanchao Bai, Zhe Wu, Jianlin Zhang, and Yongmei Huang.
\newblock Learning spatial-frequency transformer for visual object tracking.
\newblock {\em IEEE Transactions on Circuits and Systems for Video Technology}, 2023.

\bibitem{achiam2023gpt}
Josh Achiam, Steven Adler, Sandhini Agarwal, Lama Ahmad, Ilge Akkaya, Florencia~Leoni Aleman, Diogo Almeida, Janko Altenschmidt, Sam Altman, Shyamal Anadkat, et~al.
\newblock Gpt-4 technical report.
\newblock {\em arXiv preprint arXiv:2303.08774}, 2023.

\bibitem{li2024llava}
Bo~Li, Yuanhan Zhang, Dong Guo, Renrui Zhang, Feng Li, Hao Zhang, Kaichen Zhang, Peiyuan Zhang, Yanwei Li, Ziwei Liu, et~al.
\newblock Llava-onevision: Easy visual task transfer.
\newblock {\em arXiv preprint arXiv:2408.03326}, 2024.

\bibitem{wang2025internvl3}
Weiyun Wang, Zhangwei Gao, Lixin Gu, Hengjun Pu, Long Cui, Xingguang Wei, Zhaoyang Liu, Linglin Jing, Shenglong Ye, Jie Shao, et~al.
\newblock Internvl3. 5: Advancing open-source multimodal models in versatility, reasoning, and efficiency.
\newblock {\em arXiv preprint arXiv:2508.18265}, 2025.

\bibitem{wang2025unlocking}
Jingchao Wang, Zhijian Wu, Dingjiang Huang, Yefeng Zheng, and Hong Wang.
\newblock Unlocking the potential of mllms in referring expression segmentation via a light-weight mask decoder.
\newblock {\em arXiv preprint arXiv:2508.04107}, 2025.

\bibitem{cui2022mixformer}
Yutao Cui, Cheng Jiang, Limin Wang, and Gangshan Wu.
\newblock Mixformer: End-to-end tracking with iterative mixed attention.
\newblock In {\em Proceedings of the IEEE/CVF conference on computer vision and pattern recognition}, pages 13608--13618, 2022.

\bibitem{gao2022aiatrack}
Shenyuan Gao, Chunluan Zhou, Chao Ma, Xinggang Wang, and Junsong Yuan.
\newblock Aiatrack: Attention in attention for transformer visual tracking.
\newblock In {\em European Conference on Computer Vision}, pages 146--164. Springer, 2022.

\bibitem{zhou2023joint}
Li~Zhou, Zikun Zhou, Kaige Mao, and Zhenyu He.
\newblock Joint visual grounding and tracking with natural language specification.
\newblock In {\em Proceedings of the IEEE/CVF conference on computer vision and pattern recognition}, pages 23151--23160, 2023.

\bibitem{zhang2023all}
Chunhui Zhang, Xin Sun, Yiqian Yang, Li~Liu, Qiong Liu, Xi~Zhou, and Yanfeng Wang.
\newblock All in one: Exploring unified vision-language tracking with multi-modal alignment.
\newblock In {\em Proceedings of the 31st ACM International Conference on Multimedia}, pages 5552--5561, 2023.

\bibitem{zheng2023toward}
Yaozong Zheng, Bineng Zhong, Qihua Liang, Guorong Li, Rongrong Ji, and Xianxian Li.
\newblock Toward unified token learning for vision-language tracking.
\newblock {\em IEEE Transactions on Circuits and Systems for Video Technology}, 34(4):2125--2135, 2023.

\bibitem{citetracker}
Xin Li, Yuqing Huang, Zhenyu He, Yaowei Wang, Huchuan Lu, and Ming-Hsuan Yang.
\newblock Citetracker: Correlating image and text for visual tracking.
\newblock In {\em ICCV}, 2023.

\bibitem{cai2023robust}
Yidong Cai, Jie Liu, Jie Tang, and Gangshan Wu.
\newblock Robust object modeling for visual tracking.
\newblock In {\em Proceedings of the IEEE/CVF international conference on computer vision}, pages 9589--9600, 2023.

\bibitem{gao2023generalized}
Shenyuan Gao, Chunluan Zhou, and Jun Zhang.
\newblock Generalized relation modeling for transformer tracking.
\newblock In {\em Proceedings of the IEEE/CVF conference on computer vision and pattern recognition}, pages 18686--18695, 2023.

\bibitem{zheng2024odtrack}
Yaozong Zheng, Bineng Zhong, Qihua Liang, Zhiyi Mo, Shengping Zhang, and Xianxian Li.
\newblock Odtrack: Online dense temporal token learning for visual tracking.
\newblock In {\em AAAI}, 2024.

\bibitem{shi2024evptrack}
Liangtao Shi, Bineng Zhong, Qihua Liang, Ning Li, Shengping Zhang, and Xianxian Li.
\newblock Explicit visual prompts for visual object tracking.
\newblock In {\em AAAI}, 2024.

\bibitem{ma2024unifying}
Yinchao Ma, Yuyang Tang, Wenfei Yang, Tianzhu Zhang, Jinpeng Zhang, and Mengxue Kang.
\newblock Unifying visual and vision-language tracking via contrastive learning, 2024.

\bibitem{xie2024autoregressive}
Jinxia Xie, Bineng Zhong, Zhiyi Mo, Shengping Zhang, Liangtao Shi, Shuxiang Song, and Rongrong Ji.
\newblock Autoregressive queries for adaptive tracking with spatio-temporal transformers.
\newblock In {\em Proceedings of the IEEE/CVF Conference on Computer Vision and Pattern Recognition}, pages 19300--19309, 2024.

\bibitem{sutrack}
Xin Chen, Ben Kang, Wanting Geng, Jiawen Zhu, Yi~Liu, Dong Wang, and Huchuan Lu.
\newblock Sutrack: Towards simple and unified single object tracking.
\newblock 2025.

\bibitem{xu2025less}
Chenlong Xu, Bineng Zhong, Qihua Liang, Yaozong Zheng, Guorong Li, and Shuxiang Song.
\newblock Less is more: Token context-aware learning for object tracking.
\newblock In {\em Proceedings of the AAAI Conference on Artificial Intelligence}, volume~39, pages 8824--8832, 2025.

\bibitem{feng2025enhancing}
Xiaokun Feng, Dailing Zhang, Shiyu Hu, Xuchen Li, Meiqi Wu, Jing Zhang, Xiaotang Chen, and Kaiqi Huang.
\newblock Enhancing vision-language tracking by effectively converting textual cues into visual cues.
\newblock In {\em ICASSP 2025-2025 IEEE International Conference on Acoustics, Speech and Signal Processing (ICASSP)}, pages 1--5. IEEE, 2025.

\bibitem{wang2021transformer}
Ning Wang, Wengang Zhou, Jie Wang, and Houqiang Li.
\newblock Transformer meets tracker: Exploiting temporal context for robust visual tracking.
\newblock In {\em Proceedings of the IEEE/CVF conference on computer vision and pattern recognition}, pages 1571--1580, 2021.

\bibitem{zhang2021learn}
Zhipeng Zhang, Yihao Liu, Xiao Wang, Bing Li, and Weiming Hu.
\newblock Learn to match: Automatic matching network design for visual tracking.
\newblock In {\em Proceedings of the IEEE/CVF international conference on computer vision}, pages 13339--13348, 2021.

\bibitem{feng2021siamese}
Qi~Feng, Vitaly Ablavsky, Qinxun Bai, and Stan Sclaroff.
\newblock Siamese natural language tracker: Tracking by natural language descriptions with siamese trackers.
\newblock In {\em Proceedings of the IEEE/CVF conference on computer vision and pattern recognition}, pages 5851--5860, 2021.

\bibitem{guo2022divert}
Mingzhe Guo, Zhipeng Zhang, Heng Fan, and Liping Jing.
\newblock Divert more attention to vision-language tracking.
\newblock {\em Advances in Neural Information Processing Systems}, 35:4446--4460, 2022.

\bibitem{li2022cross}
Yihao Li, Jun Yu, Zhongpeng Cai, and Yuwen Pan.
\newblock Cross-modal target retrieval for tracking by natural language.
\newblock In {\em Proceedings of the IEEE/CVF Conference on Computer Vision and Pattern Recognition}, pages 4931--4940, 2022.

\bibitem{hong2024onetracker}
Lingyi Hong, Shilin Yan, Renrui Zhang, Wanyun Li, Xinyu Zhou, Pinxue Guo, Kaixun Jiang, Yiting Chen, Jinglun Li, Zhaoyu Chen, et~al.
\newblock Onetracker: Unifying visual object tracking with foundation models and efficient tuning.
\newblock In {\em Proceedings of the IEEE/CVF conference on computer vision and pattern recognition}, pages 19079--19091, 2024.

\bibitem{shao2024context}
Yanyan Shao, Shuting He, Qi~Ye, Yuchao Feng, Wenhan Luo, and Jiming Chen.
\newblock Context-aware integration of language and visual references for natural language tracking.
\newblock In {\em Proceedings of the IEEE/CVF Conference on Computer Vision and Pattern Recognition}, pages 19208--19217, 2024.

\bibitem{shi2025mamba}
Liangtao Shi, Bineng Zhong, Qihua Liang, Xiantao Hu, Zhiyi Mo, and Shuxiang Song.
\newblock Mamba adapter: Efficient multi-modal fusion for vision-language tracking.
\newblock {\em IEEE Transactions on Circuits and Systems for Video Technology}, 2025.

\bibitem{10947500}
Chengao Zong, Jie Zhao, Xin Chen, Huchuan Lu, and Dong Wang.
\newblock Learning language prompt for vision-language tracking.
\newblock {\em IEEE Transactions on Circuits and Systems for Video Technology}, 35(9):9287--9299, 2025.

\bibitem{xie2025robust}
Jinxia Xie, Bineng Zhong, Qihua Liang, Ning Li, Zhiyi Mo, and Shuxiang Song.
\newblock Robust tracking via mamba-based context-aware token learning.
\newblock In {\em Proceedings of the AAAI Conference on Artificial Intelligence}, volume~39, pages 8727--8735, 2025.

\end{thebibliography}



\end{document}